\documentclass[a4paper,twocolumn]{article}

\usepackage{jsaiac}

\usepackage{color}
\usepackage{graphicx}
\usepackage{array}
\usepackage{xcolor,amsmath}
\usepackage{tcolorbox}
\usepackage{tabularx}
\usepackage{booktabs}
\usepackage{url}

\title{Llamarine: Open-source Maritime Industry-specific\\ Large Language Model}

\address{william@aitomatic.com}

\renewcommand{\first}{\textsuperscript{1}}
\renewcommand{\second}{\textsuperscript{2}}
\newcommand{\VarSty}[1]{\textnormal{\ttfamily\color{blue!90!black}#1}\unskip}

\author{%
William Nguyen\first
\and
Hong An Phan\first
\and
Konobu Kimura\second
\and
Hitoshi Maeno\second
\and
Mika Tanaka\second
\and
Quynh Le\first
\and
William Poucher\first
\and
Christopher Nguyen\first
}

\affiliate{
\first{} Aitomatic, Inc.
\and
\second{} Furuno Electric Co., Ltd.
}

\begin{abstract}
Large Language Models (LLMs) have demonstrated substantial potential in addressing complex reasoning tasks, yet their general-purpose nature often limits their effectiveness in specialized domains such as maritime navigation. To bridge this gap, we introduce Llamarine, the first open-source LLM designed specifically for maritime navigation. Llamarine 1.0 is developed through continued pretraining and fine-tuning on a high-quality corpus comprising maritime textbooks, research publications, and web text from Wikipedia. This domain-specific training enables the model to acquire expert-level knowledge in navigational principles, collision avoidance, route optimization, and regulatory compliance. Our key contributions include (a) the curation of a comprehensive maritime dataset from authoritative sources, ensuring depth and reliability in the model's knowledge base; (b) the development of a foundational model capable of reasoning about complex navigational challenges with greater accuracy than general-purpose LLMs; and (c) the establishment of a benchmark to evaluate performance in maritime-specific decision-making tasks. Experimental results demonstrate that Llamarine outperforms both general-purpose and commercial LLMs in critical navigation-related tasks, such as trajectory planning, risk assessment, and compliance with maritime regulations. By providing an open-source foundation model trained exclusively on high-quality maritime literature, Llamarine paves the way for AI-driven advancements in maritime safety, efficiency, and operational decision-making.

\end{abstract}

\def\BibTeX{{\rm B\kern-.05em{\sc i\kern-.025em b}\kern-.08em%
 T\kern-.1667em\lower.7ex\hbox{E}\kern-.125emX}}
\def\JBibTeX{\leavevmode\lower .6ex\hbox{J}\kern-0.15em\BibTeX}
\def\LaTeXe{\LaTeX\kern.15em2$_{\textstyle\varepsilon}$}

\begin{document}

\maketitle

\section{Introduction}
\begin{figure}[!htbp]
\centering
\includegraphics[width=0.5\textwidth]{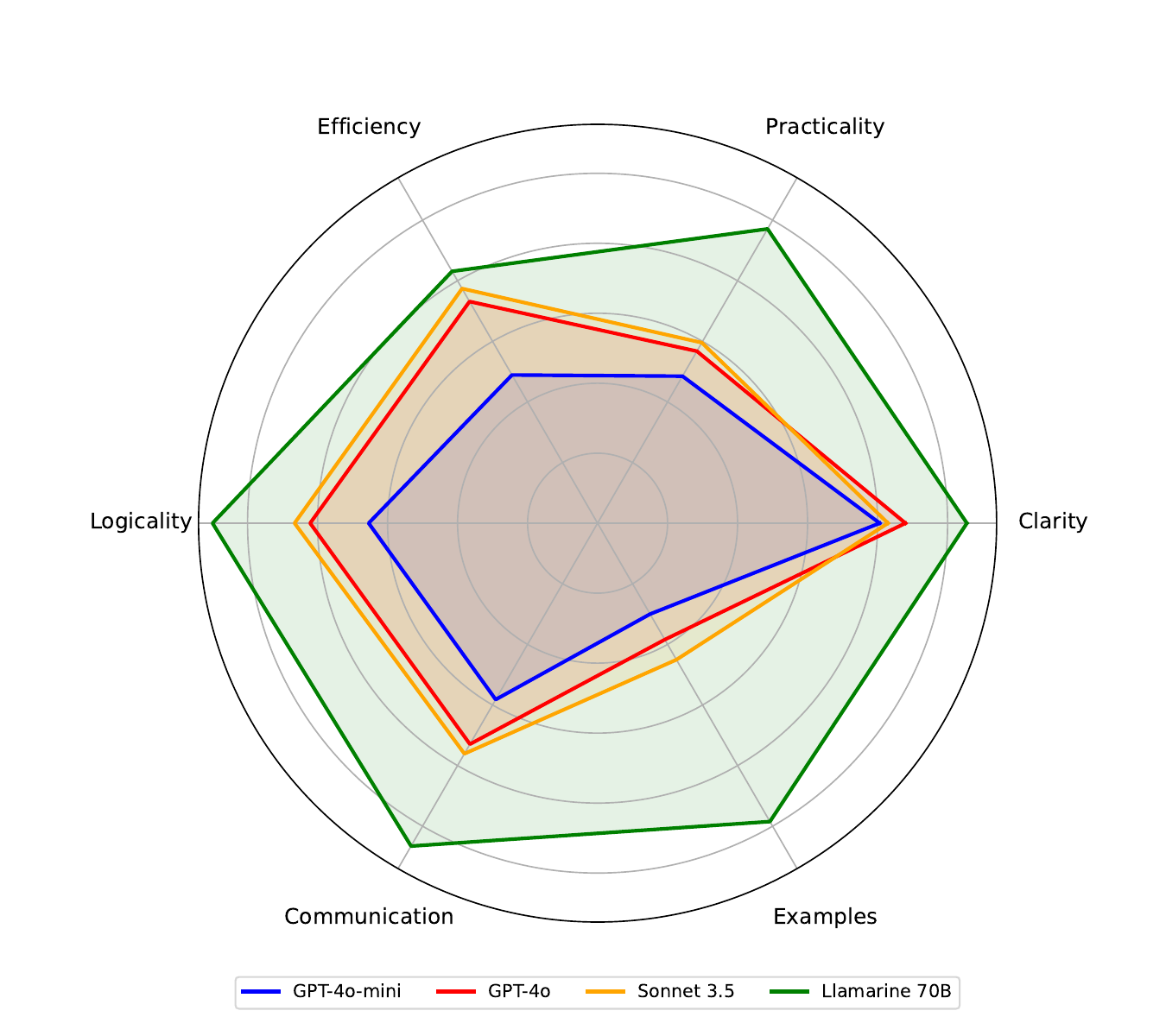} 
\caption{\textbf{Comparison of Llamarine and commercial models.}. Llamarine surpasses commercial models in every aspect, including efficiency, practicality, clarity, use of examples, expert-level communication, and logicality.}
\label{fig:products_comparison}
\end{figure}
The maritime industry, responsible for transporting approximately 80\% of global trade goods~\cite{2021ReviewOM}, operates at the intersection of extraordinary physical, human, and regulatory challenges\cite{li2023ship, zhang2022predictive, chen2024dynamics}. The vessels navigate through the most unforgiving environments on Earth, where extreme weather conditions, corrosive saltwater, and prolonged isolation create operational complexities that demand exceptional resilience from both equipment and crews~\cite{frank2019applying}. 
These challenges are compounded by the need for vessels to maintain self-sufficiency during extended voyages, often operating days away from the nearest port while performing complex three-dimensional navigation calculations that must account for dynamic weather patterns, winds, and currents~\cite{lee2022realtime}.
The human dimension of maritime operations presents equally formidable challenges. Crews work in isolation for months, navigating not only the seas but also cultural differences and language barriers within multinational teams~\cite{frank2019applying}. The industry faces a growing skill gap as experienced mariners retire~\cite{msc2023skills}, with younger generations showing a diminishing interest in seafaring careers. This knowledge drain occurs against a backdrop of increasingly complex regulatory requirements, where vessels must comply with overlapping jurisdictions and frequently updated international standards such as MARPOL and SOLAS~\cite{akpinar2021managing}.
Technological advancement in the maritime sector faces unique obstacles. The industry must manage extraordinary asset costs while attempting digital transformation with limited connectivity at sea. Port operations require intricate optimization across multiple dimensions~\cite{farzadmehr2023contemporary}, from berth allocation and crane operations to vessel queuing and environmental sustainability~\cite{mekkaoui2020way, cammin2020applications}. The emergence of autonomous vessel technologies introduces additional complexities in cybersecurity, regulatory compliance, and system integration~\cite{menges2024digital}. Furthermore, the industry faces increasing pressure to reduce the environmental impact through improved fuel management and route optimization~\cite{guzelbulut2024artificial}, all while maintaining safety standards and operational efficiency.

These multifaceted challenges create a pressing need for advanced technological solutions that can effectively manage complexity while supporting decision-making in critical maritime operations. Although artificial intelligence has shown promise in various industries~\cite{frank2019applying}, general-purpose AI solutions often do not address the unique requirements of the maritime sector~\cite{farzadmehr2023contemporary}. The specialized nature of maritime operations, combining domain-specific knowledge, safety-critical decisions, and complex regulatory compliance, demands purpose-built solutions~\cite{frank2019applying}.
This paper presents three key contributions to address these challenges:
\begin{enumerate}
\item We introduce the first open source large language foundation model focused on maritime languages, specifically designed to encode and process domain-specific maritime knowledge, including regulations, procedures, and operational expertise. This model understands the nuanced context of maritime operations, from navigation and safety protocols to compliance requirements and emergency procedures.

\item We make this model openly available to the maritime community, representing the first open-source foundation model specifically tailored for maritime applications. This contribution aims to accelerate innovation and collaboration within the industry, enabling stakeholders to build on and adapt the model for specific maritime use cases.

\item We validate our model's effectiveness through rigorous evaluation by leading maritime industry experts, demonstrating its practical utility across a range of real-world maritime scenarios and its superiority over general-purpose language models in maritime-specific tasks. The effectiveness of Llamarine has been demonstrated by outperforming generic models as shown in Figure~\ref{fig:products_comparison}.
\end{enumerate}

\section{Related Works}

Recent research has demonstrated the expanding role of Large Language Models (LLMs) \cite{ouyang2022training, openai2022chatgpt, Nguyen2024OSCaROS, Bi2024EAGLEEA, Bi2023MISARAM} across maritime applications \cite{fu2023mme, cox2023utilizing, tu2023towards, hong20233d}. Publication trend analysis indicates rapid growth in AI applications within the maritime domain \cite{xiao2024application, sanchez2019toward, li2023environmental}, spanning from shipping operations \cite{jian2022research,xu2024spatial} to traffic management \cite{liu2024data}.

Maritime LLM implementations can be deployed through various configurations, including onshore facilities, remote cloud services, and vessel-based systems, enabling real-time decision support \cite{pei2024application}. These systems have proven particularly valuable for Maritime Autonomous Surface Ships (MASSs) in processing assistance requests and generating navigation guidance \cite{wang2024kunpeng,zhang2024maritime}. Beyond basic navigation, LLMs demonstrate sophisticated capabilities in comprehending complex navigational environments and addressing long-tail issues that challenge traditional systems \cite{pei2024application,zhang2024maritime}. A foundational application involves maritime-context text identification, which forms the basis for specialized maritime Q\&A systems \cite{zhang2024maritime,wang2024kunpeng}.

The integration of LLMs has enhanced multiple aspects of maritime operations. In safety and efficiency, applications include risk assessment through natural language processing \cite{wang2024kunpeng}, vessel trajectory prediction \cite{wang2024kunpeng,drapier2024enhancing}, and traffic management through improved ETA predictions \cite{zhang2024maritime,wang2024kunpeng}. In intelligent maritime systems, LLMs facilitate smart vessel development, route optimization, and safe navigation frameworks \cite{wang2024kunpeng}. Specialized applications extend to AUV mission planning \cite{wang2024kunpeng}, automatic vessel draft reading \cite{zhang2024maritime}, and ocean science tasks \cite{wang2024kunpeng,zheng2024exploring}.

Current machine learning techniques in maritime applications show significant advancement \cite{pei2024application, GmezRuiz2023ApplicationOM, Gerakoudi2024AML}. Transformer-based architectures have emerged as particularly effective, especially in trajectory prediction through sparse augmented data representation \cite{drapier2024enhancing}. Advanced implementations of Causal Language Modeling (CLM) combined with specialized spatial indexing techniques have shown promise in maritime trajectory forecasting \cite{drapier2024enhancing}. In text processing, BERT variants demonstrate superior performance in maritime-context identification \cite{zhang2024maritime}.

However, these advanced techniques face distinct challenges. Models often struggle with domain-specific knowledge requirements \cite{zheng2023marinegpt,zheng2024exploring}, particularly in fine-grained maritime recognition tasks. Data challenges persist in developing comprehensive maritime-specific datasets \cite{zheng2023marinegpt}. Technical limitations include hallucination in advanced language models \cite{zheng2023marinegpt}, difficulties with complex maritime scenarios \cite{pei2024application}, and challenges in spatial-temporal modeling \cite{drapier2024enhancing}. Substantial computational requirements \cite{pei2024application} and limitations in current prompt engineering techniques \cite{zheng2024exploring} present additional implementation challenges.

The evolution of domain-specific LLMs has followed two main approaches: specialized pre-training and fine-tuning. Some models are pre-trained from scratch for specific domains, such as scientific research \cite{taylor2022galactica} and programming \cite{nijkamp2022codegen}. The development of instruction-following capabilities has led to numerous specialized variants, particularly in interactive applications. This trend has produced domain-adapted models across various sectors, including medicine \cite{guo2023dr, li2023cancergpt} and finance \cite{wu2023bloomberggpt, yang2023fingpt}. These specialized adaptations, supported by emerging fine-tuning frameworks, have significantly improved performance in domain-specific tasks.

\section{Llamarine: Maritime Industry Specific Large Language Model}
To develop Llamarine, a domain specific language model for maritime applications, we curated a large scale, high quality dataset tailored to the maritime industry. Our dataset comprises two main components: (1) data for continued pretraining and (2) data for supervised fine-tuning (SFT). This dual approach ensures that Llamarine not only acquires broad technical knowledge of the maritime sector but also gains specialized expertise for domain-specific tasks.
In this section, we detail our data curation process for pretraining, SFT, and benchmarking in Sections~\ref{sec:pretraining-curation}, \ref{sec:sft-curation}, and \ref{sec:eval-data-curation}, respectively. Section~\ref{sec:model-finetuning} covers model fine-tuning, while Section~\ref{sec:model-evaluation} discusses the evaluation protocol.

\subsection{Pre-training Data}
\label{sec:pretraining-curation}
To ensure Llamarine acquires a deep understanding of the maritime industry, we conducted continued pretraining using a curated dataset consisting of diverse, high-quality sources. Unlike general-purpose language models, which often lack specialized maritime knowledge, Llamarine was trained on a corpus specifically designed to cover the nuances of ship operations, maritime regulations, oceanographic data, and industry best practices.

The dataset was compiled from authoritative textbooks that provide a strong and in-depth foundational knowledge of maritime operations, regulations, and technical disciplines. These comprehensive texts serve as essential educational resources, covering international maritime regulatory frameworks, technical principles, legal case studies, and operational best practices. Foundational maritime law and compliance textbooks formed a significant portion of the corpus, offering detailed explanations of critical conventions such as SOLAS (Safety of Life at Sea), MARPOL (International Convention for the Prevention of Pollution from Ships), and STCW (Standards of Training, Certification, and Watchkeeping for Seafarers). In addition, national maritime law textbooks from agencies like the U.S. Coast Guard (USCG), the European Maritime Safety Agency (EMSA), and the UK Maritime and Coastguard Agency (MCA) ensured comprehensive coverage of legal requirements across different jurisdictions. Advanced legal texts analyzing maritime disputes, arbitration rulings, and liability cases further reinforced the dataset’s depth in maritime law. Beyond regulatory and legal aspects, the dataset included authoritative textbooks on ship engineering, navigation systems, and operational methodologies, providing rigorous technical instruction. These texts delved into ship propulsion technologies, radar and sonar applications, cargo handling systems, and vessel maintenance strategies, ensuring a solid theoretical and practical grounding. Industry-standard textbooks on maritime safety and emergency response detailed fuel efficiency optimization, distress signaling procedures, firefighting techniques, and vessel evacuation protocols, offering in-depth knowledge essential for seafarers and maritime professionals. The structured instructional content in these books provided not only theoretical insights but also applied learning, enabling a comprehensive understanding of maritime science, engineering, and operational standards.

To ensure a strong foundation in state-of-the-art maritime research, the dataset incorporated scientific and technical papers collected from arXiv, offering rigorous insights into contemporary advancements, logical reasoning, and mathematical modeling in the field. These papers provided in-depth coverage of cutting-edge topics such as AI-driven navigation, autonomous vessel control, predictive maintenance, and optimization of maritime logistics. The structured and logically precise nature of scientific publications ensured that Llamarine could interpret and generate content with a high degree of technical accuracy and coherence. Many of the collected papers presented formalized methodologies, including mathematical formulations, probabilistic models, and algorithmic frameworks that underpin modern maritime technology. Studies on fluid dynamics, structural integrity analysis, and hydrodynamic simulations contributed to a deeper understanding of vessel performance and safety. Additionally, optimization models for fuel efficiency, route planning, and risk assessment reinforced Llamarine’s ability to generate data-driven insights. By integrating research papers, the dataset not only strengthened its theoretical grounding but also ensured that Llamarine remained informed about the latest advancements in maritime science and engineering. This fusion of academic rigor and real-world applicability enabled the model to produce content that is both logically sound and technologically current.

To enhance the dataset’s coverage of general maritime knowledge, publicly available text data was sourced from Wikipedia, a widely used digital encyclopedia that provides structured and regularly updated content. Wikipedia articles serve as a valuable resource for consolidating information on maritime history, vessel operations, regulations, and industry practices. The inclusion of Wikipedia-based text data ensured that the dataset incorporated well-documented events and foundational knowledge relevant to maritime studies. A significant portion of the Wikipedia corpus included accounts of maritime accidents, offering insights into causes, consequences, and regulatory responses. These articles provided structured summaries of investigations, safety measures, and policy changes that followed maritime incidents. Additionally, Wikipedia’s extensive coverage of ship types, international agreements, and port infrastructure contributed to a broader understanding of maritime transportation and logistics. Furthermore, to support learning in oceanography, meteorology, and ship engineering, Wikipedia entries on weather conditions, ocean currents, and vessel technology were included. These resources provided an overview of environmental and technical factors affecting maritime navigation and global shipping operations. By integrating Wikipedia text data, the dataset gained factual, widely referenced, and systematically structured content, reinforcing the model’s ability to process and generate maritime-related text with clarity and contextual depth.

Once collected, the dataset underwent a multi-step processing pipeline to ensure high quality and usability. Crawled PDF files were then converted to raw text using the PyPDF library. Since the raw text often has formatting issues, we employed GPT-4o-mini for post-processing to transform the text into markdown format. This step not only corrected parsing errors but also preserved special types of information, such as tables. The final dataset for continued pretraining consisted of 117 books and 901 research papers in the domain of maritime. In this version, we primarily focus on text-based PDF data. Future work will extend to scanned documents by incorporating OCR technologies~\cite{Nguyen2024EfficientlyLL, Nguyen2021DictionaryguidedST}.

By leveraging this highly structured and domain-specific dataset, Llamarine was able to surpass general-purpose language models in maritime-specific knowledge with in depth and high quality data. The continued pretraining phase provided the necessary foundation for the model’s specialization, preparing it for fine-tuning on task-specific maritime applications.

\begin{table*}[t!]\centering
\begin{minipage}{1.0\textwidth}\vspace{0mm}    \centering
\begin{tcolorbox} 
    \centering
   
      \footnotesize
    \begin{tabular}{p{0.97\textwidth} c}
    
   \VarSty{ {\bf Theory Question:} } & \\
How does COG differ from the vessel's Heading, and why is this distinction important when navigating in complex maritime environments?
   
   \hrulefill & \\

      \VarSty{ {\bf Normal Practical Question:} } & \\
I've observed discrepancies between radar and AIS target information. What procedures should I follow to reconcile these differences and ensure accurate situational awareness?
   
   \hrulefill & \\
   
   \VarSty{ {\bf Calculation Question:} }& \\
I'm observing a celestial object from a moving vessel, which navigates at 18 knots due east. The initial sextant reading was 66°24.1' at 2200, and it's now 2230. How should I compute the object's new position, considering the vessel's movement?
    \end{tabular}
\end{tcolorbox}

\caption{\textbf{Examples of Generated Questions from Our Pipeline.} This table presents three types of questions produced by our pipeline: theoretical, practical, and calculation-based questions.}
\label{tab:question-generation-examples}
\end{minipage}
\end{table*}

\begin{figure}
    \centering
    \includegraphics[width=\columnwidth]{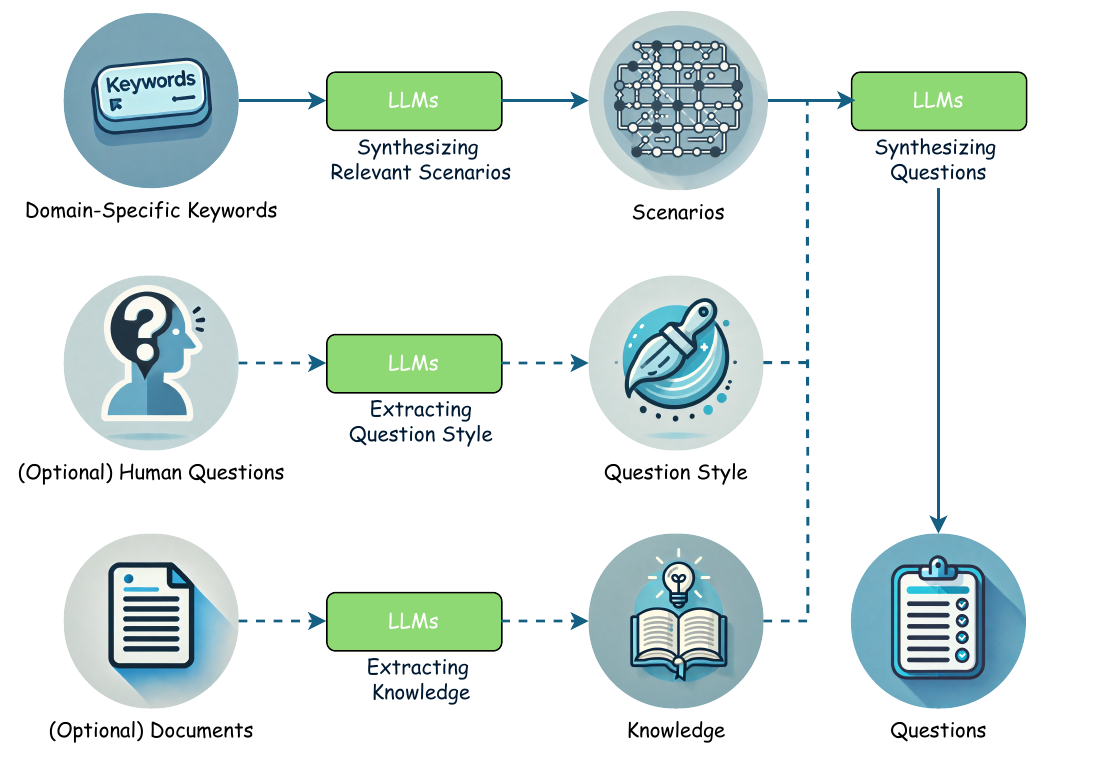}
    \caption{\textbf{Question Generation Process.} 
    The system employs LLMs in a two-step process to generate maritime-domain questions. First, it synthesizes realistic scenarios from domain-specific keywords, covering navigation, vessel operations, engineering, and regulations. Optionally, additional domain knowledge can be extracted from relevant documents and sample human questions. Finally, LLMs synthesize specific questions based on these scenarios, resulting in 56,257 questions across three categories: maritime concepts (4,852), mathematical reasoning (6,065), and operational/regulatory challenges (45,340).
    }
    \label{fig:question-synthesis}
\end{figure}

\subsection{Supervised Finetuning Data Curation}
\label{sec:sft-curation}
The supervised finetuning (SFT) dataset construction process was meticulously designed to develop a high-quality, domain-specific instruction set tailored to the maritime industry. To ensure comprehensive coverage of maritime operations, engineering, and regulatory compliance, the dataset was generated through a structured two-phase methodology: question generation and answer generation. In the first phase, a two-step approach was employed to create realistic and contextually rich questions as described in Figure~\ref{fig:question-synthesis}. Initially, GPT-4o was utilized to synthesize maritime-related scenarios derived from predefined domain knowledge and key terminology. These scenarios, which encompass areas such as navigation, vessel operations, engineering principles, and regulatory frameworks, served as contextual foundations for question formulation. We randomly add knowledge documents to a portion of questions as a way to inject domain knowledge to data. Questions that generated with documents will be also solved with equivalent documents in the answer generation phase. Additionally, to make the generated questions more natural and close with questions asked by humans, we employed a module called question style extraction. From maritime related questions crawled from internet, we extracted the style of those questions, described in text format, and randomly inject these description to question generation process. Subsequently, GPT-4o was prompted to generate specific questions based on these synthesized scenarios (mandatory), style and documents (optional), ensuring that the dataset captures real-world maritime challenges rather than abstract or generic queries. The final dataset comprises 4,852 questions explaining fundamental maritime concepts, 6,065 questions requiring mathematical reasoning for navigational and engineering calculations, and 45,340 questions addressing standard operational challenges and regulatory compliance issues. Table~\ref{tab:question-generation-examples} shows examples of these questions.

\begin{figure}
    \centering
    \includegraphics[width=\columnwidth]{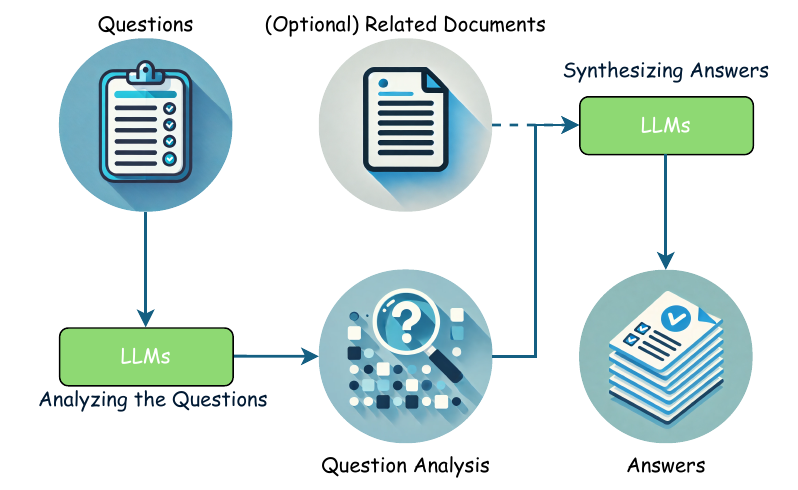}
    \caption{\textbf{Answer Generation Process.} 
    The process employs a two-step method where first, an LLM analyzes and breaks down the question into key components and reasoning paths. Second, these structured insights and optional related documents are used to generate the final answers.
    }
    \label{fig:answer-synthesis}
    \vspace{4mm}
\end{figure}

The answer generation phase followed a structured two-step approach as shown in Figure~\ref{fig:answer-synthesis}, to enhance the logical coherence, factual accuracy, and reasoning depth of the responses. In the first step, GPT-4o was prompted to analyze each question, deconstructing it into key components and identifying logical reasoning paths. This intermediate analysis provided a structured problem-solving framework, which was subsequently incorporated into a second-stage prompt, instructing GPT-4o to generate the final response based on the identified reasoning structure and equivalent documents (if any). By leveraging this chain-of-thought methodology, the model produced more accurate and logically consistent responses compared to direct answer generation. To further optimize the quality of outputs, GPT-4o was primarily used for answering conceptual and routine maritime questions, while GPT-o1-preview was employed for complex problem-solving tasks requiring advanced mathematical computations and logical deductions. This rigorous dataset development process ensures that Llamarine is equipped to address a broad spectrum of maritime challenges, ranging from fundamental theoretical knowledge to intricate multi-step problem-solving in real-world scenarios, thereby establishing a robust foundation for a maritime industry-specific LLM.

\subsection{Evaluation Data Curation}
\label{sec:eval-data-curation}
To assess the quality of our model, we introduce a novel benchmark specifically designed for the maritime domain. This benchmark comprises two distinct components: synthetic and real-world questions. The synthetic questions are generated using the same pipeline employed in the SFT training data generation process. This subset consists of 400 questions distributed across three categories: theoretical, practical, and computational problems, ensuring a diverse and comprehensive evaluation dataset. The second component includes 665 maritime-related questions sourced from the Stack Exchange forum. As these questions were posed by real users, they are inherently complex and high-quality, making them well suited to evaluate a model's ability to address real-world maritime challenges. By combining these two components, our benchmark with 1065 questions not only ensures diversity in evaluation data but also provides a rigorous assessment of how effectively models can generalize to practical, real-world scenarios in the maritime domain. We follow the evaluation method of the SemiKong~\cite{Nguyen2024SemiKongCT}, which provides a structured approach for assessing model performance across diverse question types and ensures robust evaluation results.

\subsection{Model Finetuning}
\label{sec:model-finetuning}
The curated dataset described in Section~\ref{sec:pretraining-curation} and~\ref{sec:sft-curation} was used to train our Llamarine models, following a structured pipeline that included tokenization, pre-training, fine-tuning, and post-training optimization. Initially, the text data was tokenized using Tiktoken, a Byte Pair Encoding (BPE)-based tokenizer commonly employed in NLP tasks. To enhance the model’s ability to capture sequential dependencies, we incorporated Rotary Position Embedding (RoPE) as the positional encoding mechanism. The training process began with domain-specific pre-training, where we initialized Llamarine using the Llama3.1 70B checkpoint from Meta and further trained the model on a curated corpus of maritime texts. This step was essential to infuse the model with specialized knowledge in maritime navigation, engineering, regulations, and operations, ensuring a strong foundation in the domain. We then performed SFT to refine the model’s ability to handle maritime-related tasks, such as regulatory compliance assessment, vessel routing optimization, and incident analysis, by leveraging instruction-tuned datasets.

Following pre-training and fine-tuning, we conducted post-training optimizations to enhance the model’s efficiency and suitability for deployment. To reduce computational costs without compromising performance, we applied GPTQ, a high-precision post-training quantization technique, ensuring the model remains lightweight and efficient. Furthermore, we integrated LoRA adapters and merged them with the base model to create a streamlined yet powerful version of Llamarine, tailored for real-world maritime applications. This comprehensive training and optimization pipeline ensures that Llamarine is well-equipped to support maritime professionals in decision-making, regulatory adherence, and operational efficiency within the industry.

\subsection{Model Evaluation}
\label{sec:model-evaluation}
To evaluate the performance of Llamarine, we apply an evaluation framework inspired by SemiKong, focusing on six key dimensions that ensure the model delivers clear, practical, and domain-specific responses. First, Clarity and Directness (C\&D) measures whether answers are straightforward and easy to understand, avoiding unnecessary jargon while maintaining relevance to the query. Practicality and Immediate Usability (PIU) examines how applicable the model’s recommendations are in real-world maritime contexts, emphasizing actionable insights over abstract or theoretical discussions. Efficiency and Brevity (E\&B) ensures that responses remain concise and free from redundancy while still covering all essential details.

Beyond clarity and efficiency, the evaluation also considers structural and domain-specific aspects. Logical Flow and Coherence (LFC) assesses whether responses are well-organized and logically structured, making information easy to follow. Expert-to-Expert Communication (EEC) focuses on whether the model provides insights suitable for maritime professionals, maintaining technical depth without unnecessary simplifications. Finally, Use of Examples and Specificity (UES) evaluates the appropriateness of examples and technical details, ensuring they add value without overcomplicating explanations. By incorporating these six dimensions, we ensure that Llamarine produces responses that are not only informative and accurate but also tailored to the needs of the maritime industry.

\begin{table*}[t!]\centering
\begin{minipage}{1.0\textwidth}\vspace{0mm}    \centering
\begin{tcolorbox} 
    \centering
   
      \footnotesize
    \begin{tabular}{p{0.97\textwidth} c}
   \VarSty{ {\bf Question:} } &\\
    A vessel on my port side coming to a crossing situation on my radar. SOG is being used. What should I do?

    \hrulefill & \\
    
   \VarSty{ {\bf GPT-4o's answer:} } & \\
In a crossing situation where a vessel is on your port side and Speed Over Ground (SOG) is being used on your radar, the correct action depends on the International Regulations for Preventing Collisions at Sea (COLREGs). According to Rule 15 (Crossing Situation) of COLREGs:

You are the stand-on vessel because the other vessel is on your port side.
The vessel on your port side is the give-way vessel and must take early and substantial action to avoid a collision.

As the stand-on vessel, you must maintain your course and speed unless it becomes apparent that the give-way vessel is not taking appropriate action.
If the give-way vessel does not take action and a collision risk develops, Rule 17 (Action by Stand-on Vessel) allows you to take action to avoid the collision.
Avoid turning to port for a vessel on your port side, as this could increase the risk of collision.

\textbf{Recommended Action:}

Maintain your course and speed initially.

Monitor the other vessel closely on radar and visually.

Communicate if necessary via VHF radio to clarify intentions.

Be prepared to act if the give-way vessel does not comply, by altering course to starboard or slowing down if required to avoid a close-quarters situation.

Would you like further clarification on radar interpretation in collision avoidance scenarios?
   
   \hrulefill & \\

      \VarSty{ {\bf Sonnet 3.5's answer:} } & \\
As a vessel approaching from your port side in a crossing situation, you are the stand-on vessel according to the International Regulations for Preventing Collisions at Sea (COLREGS) - specifically Rule 15. However, this comes with important responsibilities:

Initially maintain your course and speed as the stand-on vessel.
Monitor the other vessel carefully using all available means (radar, visual observation, AIS if available). The fact that you're using SOG (Speed Over Ground) from radar is good, but remember to cross-reference with visual observation if possible.

Take early action if it becomes clear the give-way vessel (the one on your port side) is not taking appropriate action to keep clear. According to Rule 17, if you find yourself in a situation where collision cannot be avoided by the give-way vessel's actions alone, you must take action.

If you need to take action, make it clear and decisive enough to be obvious to the other vessel. Avoid a series of small alterations.

Consider giving sound signals (five short blasts) if you're in doubt about the other vessel's intentions.

Are you able to determine if the other vessel is taking appropriate action to stay clear? This would help me provide more specific guidance for your situation.
   
   \hrulefill & \\
   
   \VarSty{ {\bf Llamarine's answer:} }& \\

Given the situation where a vessel is on your port side and approaching a crossing situation, as the stand-on vessel, your primary responsibility is to maintain your course and speed. However, this does not mean you should be complacent; rather, you must remain vigilant and prepared to take action if necessary.

\textbf{1. Maintain Course and Speed:}
   
   - Continue on your current course and speed, ensuring that you do not alter your trajectory in a way that could be perceived as an attempt to give way. This is crucial because altering course or speed could be misinterpreted by the give-way vessel.
   
   - Example: If your current course is 090° and speed is 12 knots, maintain these parameters.

\textbf{2. Monitor the Situation Closely:}
   
   - Continuously monitor the give-way vessel's movements and intentions. Use your radar and AIS (Automatic Identification System) to track its speed over ground (SOG), course over ground (COG), and CPA (Closest Point of Approach).
   
   - Example: If the CPA is less than 0.5 nautical miles, consider it a close quarters situation that warrants extra attention.

\textbf{3. Assess the Give-way Vessel's Actions:}
   
   - Observe whether the give-way vessel is taking appropriate action to avoid a collision. It should alter its course and/or speed to ensure a safe passing distance.
   
   - Example: If the give-way vessel alters its course to starboard, increasing the CPA to more than 1 nautical mile, it indicates effective action.

\textbf{4. Prepare for Emergency Action:}
   
   - Despite being the stand-on vessel, be prepared to take evasive action if it becomes clear that the give-way vessel is not taking appropriate action to avoid a collision.
   
   - Example: If the CPA suddenly decreases to less than 0.2 nautical miles and the give-way vessel shows no signs of altering its course or speed, consider taking immediate action such as reducing speed or altering course to starboard.

\textbf{5. Communicate if Necessary:}
   
   - If there is any doubt about the intentions of the give-way vessel or if you decide to take evasive action, use VHF radio to communicate your intentions clearly.
   
   - Example: "This is [Your Vessel Name], altering course to starboard to avoid collision. Please confirm your intentions."

By following these steps, you ensure that you are fulfilling your obligations as the stand-on vessel while also being prepared to adapt to any unexpected changes in the situation.

    \end{tabular}
\end{tcolorbox}

\caption{\textbf{Qualitative comparison of Llamarine and commercial models GPT-4o and Sonnet 3.5.} Llamarine delivers practical answers, emphasizing useful knowledge and real-world examples to help users make informed decisions, rather than merely repeating basic theory. Its solutions follow a logical sequence that aligns with real-world scenarios.}
    \label{tab:qualitative-comparison}
\end{minipage}
\end{table*}

\begin{table*}[h!]
\centering
\vspace{-4mm}
\caption{Compare with open source models}
\begin{tabularx}{\textwidth}{l *{7}{>{\centering\arraybackslash}X}}
\toprule
\textbf{Model} & \textbf{C\&D} & \textbf{PIU} & \textbf{E\&B} & \textbf{LFC} & \textbf{EEC} & \textbf{UES} & \textbf{Total} \\ \midrule
Llama 3.1 8B & 3.63 & 3.28 & 3.24 & 3.55 & 3.50 & 3.19 & 20.39 \\ 
Llama 3.1 70B & 3.95 & 3.54 & 3.34 & 3.86 & 3.69 & 3.44 & 21.82 \\ 
Llama 3.3 70B & 4.03 & 3.65 & 3.71 & 4.07 & 3.93 & 3.55 & 23.00 \\ 
\midrule
Llamarine 70B & \textbf{4.27} & \textbf{4.05} & \textbf{4.02} & \textbf{4.37} & \textbf{4.26} & \textbf{4.09} & \textbf{25.07} \\ 
\bottomrule
\end{tabularx}
\label{tab:open_source_models}
\end{table*}

\begin{table*}[h!]
\centering
\caption{Compare with commercial products}
\begin{tabularx}{\textwidth}{l *{7}{>{\centering\arraybackslash}X}}
\toprule
\textbf{Model} & \textbf{C\&D} & \textbf{PIU} & \textbf{E\&B} & \textbf{LFC} & \textbf{EEC} & \textbf{UES} & \textbf{Total} \\ \midrule
GPT-4o-mini & 3.83 & 3.33 & 3.33 & 3.59 & 3.48 & 3.04 & 20.60 \\ 
GPT-4o & 3.94 & 3.46 & 3.71 & 3.85 & 3.71 & 3.17 & 21.85 \\ 
Sonnet 3.5 & 3.86 & 3.50 & 3.78 & 3.92 & 3.76 & 3.27 & 22.08 \\ 
\midrule
Llamarine 70B & \textbf{4.21} & \textbf{4.08} & \textbf{3.86} & \textbf{4.28} & \textbf{4.23} & \textbf{4.10} & \textbf{24.76} \\ 
\bottomrule
\vspace{-1mm}
\end{tabularx}
\label{tab:commercial_products_comparison}
\vspace{-5mm}
\end{table*}

\section{Experiments}
In this section, we present experimental comparisons between Llamarine and both open-source and commercial models. For open-source benchmarks, we evaluate Llamarine against LLaMA 3.1 8B, LLaMA 3.1 70B, and LLaMA 3.3 70B, with the latter considered to achieve performance comparable to LLaMA 3.1 405B. For commercial models, we compare Llamarine with three widely used systems: GPT-4o-mini, GPT-4o, and Sonnet 3.5. Additionally, GPT-4o was employed as the evaluation tool to assess all six aspects outlined in Section~\ref{sec:model-evaluation}.

\subsection{Implementation}
Llamarine was trained using 8xA100 80GB GPUs from Llama3.1 70B, adhering to best practices outlined in the Transformers HuggingFace framework, HuggingFace Accelerator, and the LLaMA-Factory library for LLM fine-tuning. The pretraining and SFT phases employed a batch size of 3, gradient accumulation steps of 3, and a learning rate of 1.0e-5. Training was conducted on 50 million (0.05B) parameters over 2 epochs using QLoRA, with a cosine learning rate scheduler and a warm-up ratio of 0.15. To enhance computational efficiency, NF4 quantization was utilized for parameter representation. Additionally, 20\% of the dataset was allocated for validation to assess model performance.

\subsection{Experiment Results}
\label{sec:experiment-results}

Table~\ref{tab:open_source_models} presents the experimental results comparing open-source models and Llamarine using our proposed benchmark. The results indicate that LLaMA 3.3 70B outperforms LLaMA 3.1 70B across all evaluated aspects while maintaining the same computational resource requirements. Notably, although Llamarine was trained based on LLaMA 3.1 70B, it significantly outperformed LLaMA 3.3 70B across all evaluation criteria, including directness, practicality, efficiency, logicality, expert communication, and use of examples. It also means that Llamarine will potentially be enhanced in the future when we train Llamarine with stronger foundation models. These findings demonstrate the enhanced comprehensiveness and effectiveness of Llamarine.

To demonstrate that our approach effectively produces a domain-specific model for the maritime sector that outperforms general-purpose models, it is essential to compare Llamarine against proprietary systems. In this experiment, we evaluate Llamarine against some of the most advanced and widely used models available today, including GPT-4o-mini, GPT-4o, and Sonnet 3.5. A common limitation of these proprietary models is their restricted accessibility, as they can only be utilized through the provider's platform or API, raising concerns about data security and control.
As shown in Table~\ref{tab:commercial_products_comparison}, our benchmark results indicate that GPT-4o and Sonnet 3.5 achieve comparable performance across all evaluation metrics, both outperforming the lighter GPT-4o-mini model. However, when compared to Llamarine, our model demonstrates a substantial performance advantage over both GPT-4o and Sonnet 3.5 across all evaluation criteria, with particularly strong improvements in practicality, expert communication, and the use of examples. This suggests that Llamarine minimizes the repetition of fundamental theoretical concepts and instead provides precise, domain-specific guidance that enables users to address real-world problems more effectively. Such characteristics align closely with the expectations and needs of professionals working within the maritime industry.

\subsection{Ablation Study}

% To be able to interpret the quatitative experiment results in section~\ref{sec:experiment-results}, it will be easier to visualize if we analyze the results qualitatively. Table~\ref{tab:qualitative-comparison} shows the comparison when Llamarine, GPT-4o, and Sonnet 3.5 answer the same questions. GPT-4o and Sonnet 3.5 tend to focus on repeating basic knowledge, in this case COLREGs, and provide recommendations that are general and lacking in specificity. This will make it difficult for users to immediately apply these tips into practice. In contrast, Llamarine responds directly to the current situation, identifies the problem, and provides a step-by-step solution with examples for each step to make it easier for users to implement the solution in practice. These results are consistent with the original hypothesis that generic models lack domain specific knowledge, thereby making it impossible to provide answers with deep expertise. By training the model with domain specific text and fine-tuning, Llamarine can gain in-depth domain knowledge and thereby provide high-quality answers.

To better interpret the quantitative results in Section~\ref{sec:experiment-results}, we present a qualitative comparison in Table~\ref{tab:qualitative-comparison}. GPT-4o and Sonnet 3.5 primarily restate COLREGs, providing general recommendations that, while correct, lack the specificity needed for real-world decision-making. Their responses focus on theoretical guidance rather than actionable strategies, making them less practical in operational settings.
Llamarine, in contrast, delivers structured, context-aware responses. It not only acknowledges regulatory requirements but also breaks down the situation into clear, actionable steps. By incorporating real-world considerations such as Closest Point of Approach (CPA) thresholds, monitoring strategies, and explicit decision-making criteria, Llamarine ensures that its guidance is both relevant and immediately applicable. The use of concrete examples further enhances clarity, enabling mariners to implement solutions effectively.
These findings support our hypothesis that general-purpose models, despite their broad knowledge, lack the depth required for specialized domains. Without targeted fine-tuning, they struggle to provide expert-level insights. By contrast, Llamarine’s domain-specific training enables it to bridge this gap, offering responses that align with professional maritime standards and operational best practices. This underscores the importance of domain adaptation in improving the practical utility of language models for specialized fields.

\vspace{-3.5mm}
\section{Conclusion}

In this paper, we introduce Llamarine, the first foundation model specialized for maritime navigation, available in a 70B version. Additionally, we have released a large-scale dataset tailored for maritime applications, encompassing both pretraining and fine-tuning data. Our Llamarine models have achieved state-of-the-art performance, surpassing open-source foundation models and outperforming commercial products in expert-level navigation guidance. However, Llamarine represents just the initial step, with substantial opportunities for further development. First, we can expand its capabilities beyond core navigation tasks, making AI-driven maritime solutions more comprehensive across vessel operations, route optimization, and safety management. Secondly, our method can be adapted to other maritime domains, such as logistics and environmental monitoring, thereby advancing AI-driven innovations across the broader maritime industry.

\section*{Acknowledgments}

We would like to express our gratitude to the AI Alliance (\url{https://thealliance.ai}) for providing the impetus, resources, and platform for this work, and for collaboration in open science. We also extend our thanks to the member organizations of the AI Alliance, their researchers, and engineers for their valuable contributions to this study, including Anthony Annunziata (IBM Research), Sean Hughes (ServiceNow), Phong Nguyen (FPT Software, AI Center), Noritaka Yokomori (Tokyo Electron). Their expertise, insights, and collaborative spirit have been instrumental in advancing our research. 

%% Include your bibliography here
\bibliographystyle{plain}  % Style for references
\bibliography{references}  % .bib file without the extension

\end{document}